# Multi-strategy Collaborative Optimized YOLOv5s and its Application in Distance Estimation

Zijian Shen
The University of Hong Kong
Hong Kong, China

Zhenping Mu
The Hong Kong University of Science and Technology
Hong Kong, China

Xiangxiang Li*
The Hong Kong University of Science and Technology
Hong Kong, China
xliio@connect.ust.hk

*Abstract*—The increasing accident rate brought about by the explosive growth of automobiles has made the research on active safety systems of automobiles increasingly important. The importance of improving the accuracy of vehicle target detection is self-evident. To achieve the goals of vehicle detection and distance estimation and provide safety warnings, a Distance Estimation Safety Warning System (DESWS) based on a new neural network model (YOLOv5s-SE) by replacing the IoU with DIoU, embedding SE attention module, and a distance estimation method through using the principle of similar triangles was proposed. In addition, a method that can give safety suggestions based on the estimated distance using nonparametric testing was presented in this work. Through the simulation experiment, it was verified that the mAP was improved by 5.5% and the purpose of giving safety suggestions based on the estimated distance information can be achieved.

## I. Introduction

The explosive growth of vehicles such as automobiles in recent years has not only brought convenience to people's lives but also posed huge challenges to environmental protection, driving safety and pedestrian protection [1-3]. The increasing number of car accidents and fatalities on roads worldwide has caused great demand for enhanced vehicle safety system, such as the advanced driver assistance systems (ADAS) [4, 5]. The implementation of ADAS highly depends on the computer vision, which is used for real-time perception and understanding of the surrounding environment. As a key technology in the field of computer vision, target detection has shown great immense potential in helping ADAS to achieve this goal, which will finally reduce the accidents on the road [6, 7]. However, there are still many areas for improvement when applying the current target detection technology to ADAS or autonomous driving system due to the high scene complexity [8]. For instance, a changeable background, changeable light conditions and fast detection speed are needed [9].

To better improve the effectiveness of vehicle target detection, the network models based on the YOLO family are widely used. Sang et. al. proposed a vehicle detection model with a mean mAP of 94.78% based on improved YOLOv2 [10]. Later a nighttime vehicle detection model based on YOLOv3 was introduced to solve the problem of poor nighttime recognition performance [11]. Through this method, the mean mAP was still relatively high (93.66%) during the nighttime [11]. Then, a refined YOLOv4 model was proposed for higher accuracy and faster speed in real-time detection [12]. In recent studies, a network model based on YOLOv5 was used to reduce the computational complexity [13, 14]. Unfortunately, the computational complexity of YOLOv5 is still high and the overall detection precision is not high enough.

In this paper, we proposed a Distance Estimation Safety Warning System (DESWS), which uses the smaller model (YOLOv5s) as the basic model in order to decrease the computational complexity. Meanwhile, an improved method based on conventional IoU (Intersection over Union) loss named DIoU (Distance-IoU) loss, which focuses both on the overlapping areas and the distance between the target box and the predicted box [15], was considered to improve the object detection accuracy. To further improve the detection precision of the network model, we also proposed embedding the Squeeze-and-Excitation (SE) attention module [16] into our model. By doing all of this, a refined network model YOLOv5s-SE based on YOLOv5s was introduced in this paper. For further application of the improved model, a distance estimation method and a nonparametric testing and warning method were proposed to predict the distance and give safety suggestions based on the estimated distance. The structural flow chart of this system is shown in Fig. 1.

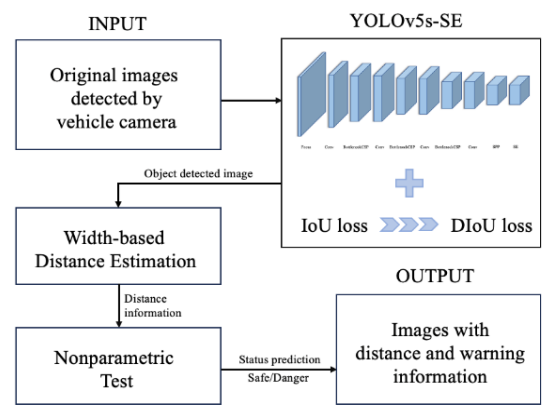

Fig. 1. The structure flow chart of DESWS

## II. MODEL

The Distance Estimation Safety Warning System (DESWS) consists primarily of two interconnected modules: object detection and distance estimation. In a real-world autonomous driving scenario, onboard cameras continuously capture images of the surrounding environment in front of the vehicle. The object detection module then identifies and classifies potential obstacles and targets in each frame using a deep learning model like YOLOv5s. The distance estimation module calculates the approximate distance to each detected object. This module can also give a warning to some objects that are too close.

### A. Object Detection Module

The object detection module utilizes a state-of-the-art deep convolutional neural network, YOLOv5s, to detect vehicles and pedestrians in the image data of the camera. YOLOv5s is optimized for real-time performance, achieving high accuracy (mAP of 51.5%) at 55 frames per second. This allows for rapid object detection even on computationally constrained systems.

YOLOv5s is a highly optimized deep convolutional neural network for real-time object detection. Building upon the successful YOLOv5 architecture, the 's' variant makes several improvements to optimize for speed and latency, making it ideally suited for applications where quick inference is a priority. These features fit well with the needs of our DESWS. Hence, object detection module applies it to ensure both high accuracy and quick response.

### B. Distance Estimation Module

The distance estimation module predicts inter-vehicle distance based on width of the target object[18]. Considering that the actual widths of different target objects vary greatly in our study scenario, we preset different actual widths for different objects, which is called adaptive width-based distance estimation method. The main idea of the algorithm is simple geometric principles, as it is shown in Fig. 2.

$D_{img}$ is the focal length of the camera, $w_{img}$ is the width of the target vehicle (in pixel unit) in the image, $D_{obj}$ is the real distance from camera to target vehicle, $w_{obj}$ is the real width of the target vehicle, according to the principle of similar triangles,

$$\frac{D_{img}}{D_{obj}} = \frac{w_{img}}{w_{obj}} \quad (1)$$

for a given camera, the focus length of it is fixed as $f$, so the $D_{obj}$ could be given as,

$$D_{obj} = \frac{w_{img}}{w_{obj}} f \quad (2)$$

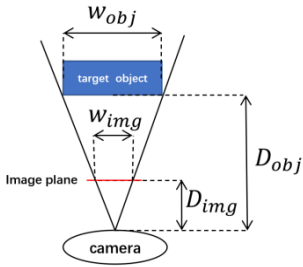

Fig. 2. The principle of adaptive width-based distance estimation

We can notice that this algorithm needs real width of the target object. Since we cannot know exactly the actual width of each object, we divide the target objects into several categories, including person, bicycle, car, motorcycle, bus, and truck, each of which has its own predetermined actual width.

Together, these two modules allow DESWS to provide critical early warnings for too-closed objects. By this method, DESWS aims to give drivers enough time to evaluate the current traveling conditions and respond to potential dangers.

## III. MULTI-STRATEGY COLLABORATIVE OPTIMIZED YOLOV5S

First, we replace the traditional IoU loss function in the original YOLOv5 framework with DIoU loss. DIoU Loss, also known as Distance-IoU Loss, is an improved method based on IoU Loss. When calculating the distance between the target box and the predicted box, DIoU Loss introduces additional information, which better describes the differences between them and makes the trained model more accurate in detection tasks. The basic idea behind DIoU Loss is to include the distance information between the target box and the forecast box while computing IoU Loss, so that the computation of Loss takes into consideration not only the overlap between them, but also the distance disparities between them. The computing formular of DIoU loss is shown in equation (3-8).

$$\text{DIou Loss} = 1 - \text{IoU} + \frac{d_c^2}{diag^2} \quad (3)$$

$$\text{Iou} = \frac{A_{overlap}}{A_{pred} + A_{target} - A_{overlap}} \quad (4)$$

$$d_c = \|C_1 - C_2\|^2 \quad (5)$$

$$diag^2 = w_{union}^2 + h_{union}^2 \quad (6)$$

$$w_{union} = \max(x_{max1}, x_{max2}) - \min(x_{min1}, x_{min2}) \quad (7)$$

$$h_{union} = \max(y_{max1}, y_{max2}) - \min(y_{min1}, y_{min2}) \quad (8)$$

Where $A_{overlap}$ represents the overlapping area between the predicted box and the true box, $A_{pred}$ represents the area of the predicted box, $A_{target}$ represents the area of the true box, $d_c$ is the Euclidean distance between the centers of the boxes, diag represents the diagonal length of the minimum bounding rectangle, $w_{union}$ and $h_{union}$ correspond to the width and height of the minimum bounding rectangle, $x_{min1}, y_{min1}, x_{max1}, y_{max1}$ represent the coordinates of the true box and $x_{min2}, y_{min2}, x_{max2}, y_{max2}$ represent the coordinates of the true box.

Secondly, the Squeeze-and-Excitation (SE) attention module [16] was embedding into the YOLOv5s network by directing adding the SE module at the end of the backbone, like shown in Fig. 3.

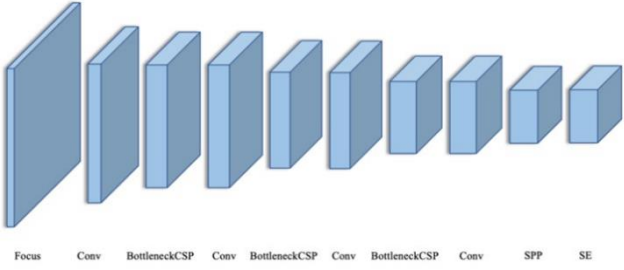

Fig. 3. The backbone structure after embedding SE attention module

## IV. EXPERIMENTAL

### A. Experimental Environment

The training dataset used in this paper is COCO dataset contains 128 images from the training set and 70 images from the validation and test sets. The labels includes 6 types of objects that are potentially dangerous to driving safety, person, bicycle, car, motorcycle, bus, truck in our application scenario.

The processor used was Intel(R) Core(TM) i9-12900HK @ 2.50GHz. The memory of it is 64.0GB (63.7GB available) and GPU is GTX 3080-ti.

### B. Results for Target Detection

The training network epochs, batch-size and image size were set to 100, 16 and [640, 640], respectively. The training process of YOLOv5s with DIoU loss is shown in Fig. 4. In Fig. 5 the training process of the final network YOLOv5s-SE, which is by replacing the IoU loss of the YOLOv5s with DIoU loss and adding the Squeeze-and-Excitation (SE) attention module, is shown.

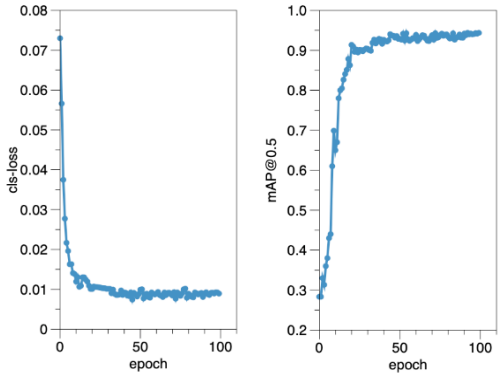

Fig. 4. Cls_loss and mAP@0.5 curve for YOLOv5s with DIoU loss

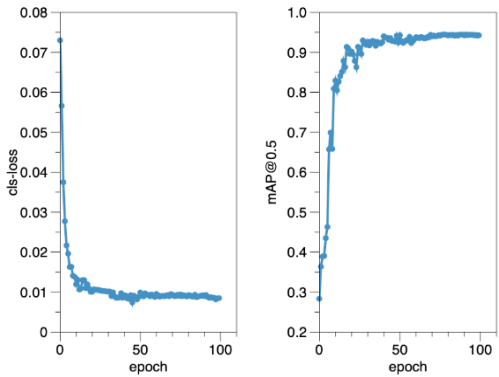

Fig. 5. Cls_loss and mAP@0.5 curve for YOLOv5s-SE

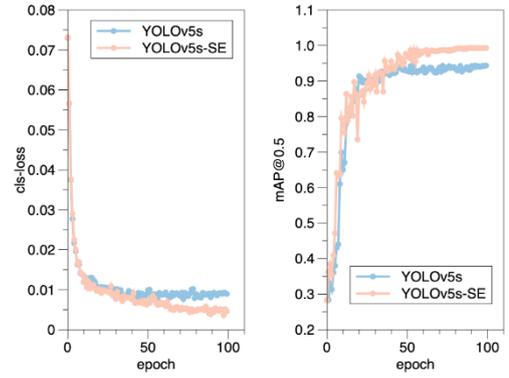

Fig. 6. The training results of YOLOv5s and YOLOv5s-SE

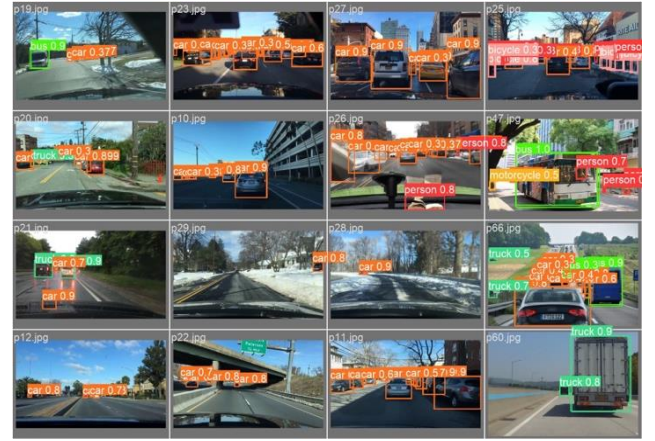

Fig. 7. The detection results in validation process

Obviously, the new network structure through replacing the IoU with DIoU and embedding SE attention module can work and by comparing the results of network with IoU loss and DIoU loss, the mAP@0.5 for YOLOv5s with DIoU loss was slightly higher and cls_loss of it was declining steadily (Fig. 4). When the SE attention module has been embedded, the network can also work steadily, the training process is shown in Fig. 5. Although there were some fluctuations in training mAP@0.5, the overall trend was good and steadily increasing. With the embedding of SE attention module, the mAP@0.5 was greatly improved. To show this, a comparison experiment between the initial YOLOv5s model and our new YOLOv5s-SE was done, the results are shown in Fig. 6. Note that cls_loss is used to supervise category classification and calculate whether the anchor box and corresponding calibration classification are correct, mAP@0.5 stands for the mAP@0.5 value under the standard of IoU>=0.5 [17]. Higher mAP@0.5 was expected because the higher mAP is, more accuracy the model will be. The results of mAP@0.5 for three models based on the training and validation set are shown in TABLE I. The detection results of validation are shown in Fig. 7.

From TABLE I it can be seen that the new network model YOLOv5s-SE does work successfully in all train, validation and test processes. Besides, the YOLOv5s-SE showed a higher mAP both in train, validation and test. More specifically, the mAP of training has increased 5.5% and in test it has been improved by 6.5%, which satisfied the application scenario and the accuracy expectations.

TABLE I. MAP@0.5 DATA

|  | With IoU | With DIoU | YOLOv5s-SE |
|---|---|---|---|
| Train | 0.944 | 0.945 | 0.993 |
| Validation | 0.944 | 0.945 | 0.993 |
| Test | 0.903 | 0.936 | 0.962 |

*C. Results for Distance Estimation*

Accurate distance estimation in conditions of lack of light is nontrivial, because in which restricted vision brings greater risks to driving safety. Our proposed DESWS shown robust performance even in dark or rainy environments, as the result in Fig. 8, showing that our algorithm could estimate the distance accurately and predict the potential dangerous of the surround pedestrians and vehicles.

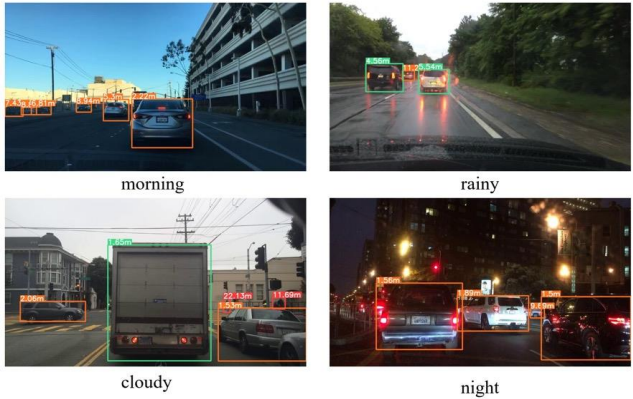

Fig. 8. The detection results in validation process

*D. Results for Distance Estimation*

According to the detected objects and their estimated distance, the module will compare it with a threshold defined by nonparametric test [19].

We organize the distance data obtained and set different thresholds. As it is shown in Table II, including the threshold, dangerous distance and the safe distance.

TABLE II. NON-PARAMETRIC TEST DATA

| Threshold | Dangerous Distance | Safe Distance |
|---|---|---|
| 3 | 10 | 8 |
| 3.5 | 10 | 8 |
| 4 | 10 | 8 |
| 4.5 | 10 | 8 |
| 5 | 9 | 9 |
| 5.5 | 8 | 10 |
| 6 | 7 | 11 |
| 6.5 | 7 | 11 |
| 7 | 6 | 12 |

Non-parametric test was conducted on the above data, and the results were shown in the Fig. 9. Non-parametric test is used to study the difference of threshold values for risk and safety. It can be seen from the table above that different threshold samples do not show significant effects on risk and safety (p> 0.05), which means that all samples with different threshold values show consistency for risks, but there is no difference. Therefore, 6 is selected as the threshold. If the value is greater than 6, it is safe and less than 6 is dangerous. The result we combine the result of non-parametric test and distance estimation, we could predict if front target object is dangerous or not, which is shown in Fig. 10.

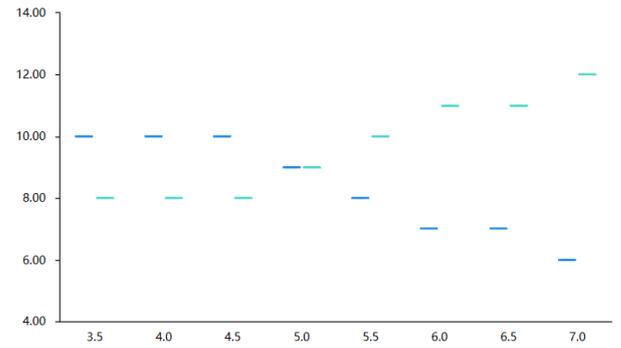

Fig. 9. The result of non-parametric test

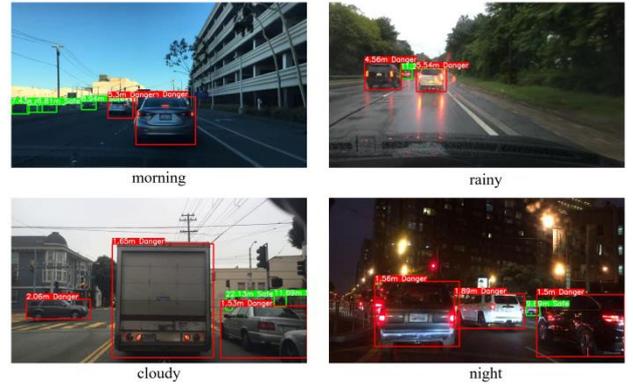

Fig. 10. The prediction of state of target objects

## V. CONCLUSION

This paper proposed DESWS, Distance Estimation Safety Warning System, to solve the problem of ensuring safe driving by distance estimation. The solution of object detection is based on YOLOv5s-SE and solution of distance estimation is based on objects 'width shown excellent performance on COCO dataset and have noticeable improvement compared to original algorithm. These methods present robust performance of ensuring driving safety to advance the state-of-the-art in vehicle safety system even in some dim environment. The further research of this paper will focus on improving the sensitivity and real time response performance of current distance estimation solution, which allows it to be better used in vehicle safety system.